\documentclass[letterpaper, 10 pt, journal, twoside]{IEEEtran}

\IEEEoverridecommandlockouts    

\usepackage[T1]{fontenc}

\usepackage{graphicx} % for pdf, bitmapped graphics files
\usepackage{amsmath} % assumes amsmath package installed
\usepackage{amsfonts}       % blackboard math symbols
\usepackage{nicefrac}       % compact symbols for 1/2, etc.
\usepackage{microtype}      % microtypography
\usepackage{subfig}			% per usare le subfigure
\usepackage{tikz}
\usepackage{adjustbox}
\usepackage{booktabs}	% toprule midrule bottomrule
\usepackage{multirow}
\usepackage{url}
\usepackage{amsthm}
\usepackage{listings}
\usepackage[linesnumbered,ruled,vlined]{algorithm2e}
\usepackage{xcolor}
\usepackage{comment}

\usepackage{hyperref}

\usepackage{pgfplots}
\pgfplotsset{compat=1.18}
\usepackage{xfrac}
\usepackage[sort&compress, numbers]{natbib}
\usepackage{flushend}
\definecolor{blu}{rgb}{0, 0,0.7}
\definecolor{verde}{rgb}{0, 0.7,0}
\definecolor{nero}{rgb}{0, 0,0}
\definecolor{rosso}{rgb}{0.7, 0,0}
\definecolor{grigio}{rgb}{0.3, 0.3,0.3}
\definecolor{bianco}{rgb}{1,1,1}
\definecolor{verdescuro}{rgb}{0.298, 0.6, 0} 
\definecolor{rossoscuro}{rgb}{0.6, 0,0}

\definecolor{black}{rgb}{0, 0, 0}
\definecolor{red}{rgb}{0.9, 0, 0}
\definecolor{green}{rgb}{0, 0.6, 0}
\definecolor{blue}{rgb}{0, 0, 0.9}
\definecolor{grey}{rgb}{0.52, 0.52, 0.51}

\newcommand{\fillbox}[3]% #1=width, #2=height, #3=filename
{\bgroup
  \dimen1=#1\relax% store width into register
  \dimen2=#2\relax% store height into register
  \sbox0{\includegraphics[width=#1]{#3}}%
  \ifdim\ht0>\dimen2
    \dimen0=\dimexpr \ht0-\dimen2\relax
    \adjustbox{clip=true,trim=0pt 0.5\dimen0 0pt 0.5\dimen0}{\usebox0}%
  \else
    \sbox0{\includegraphics[height=#2]{#3}}%
    \ifdim\wd0>\dimen1
      \dimen0=\dimexpr \wd0-\dimen1\relax
      \adjustbox{clip=true,trim=0.5\dimen0 0pt 0.5\dimen0 0pt}{\usebox0}%
    \else
      \usebox0
    \fi
  \fi
\egroup}

\DeclareMathOperator*{\argmax}{argmax}

\newcommand{\stateH}{x_h}
\newcommand{\stateR}{x_r}
\newcommand{\goalH}{g_h}
\newcommand{\goalR}{g_r}
\newcommand{\actionR}{u_r}

\newcommand{\StateH}{X_h}
\newcommand{\StateR}{X_r}
\newcommand{\GoalH}{G_h}

\newcommand{\ActionR}{U_r}
\newcommand{\GoalHSet}{\mathcal{G}_h}
\newcommand{\GoalHSetI}[1]{\mathcal{G}_{h,#1}}

\title{Learning-Based Safety-Aware Task Scheduling for Efficient Human-Robot Collaboration}

\author{M. Faroni, A. Spanò, A. M. Zanchettin, P. Rocco
\thanks{This study was partially carried out within the MICS (Made in Italy – Circular and Sustainable) Extended Partnership and received funding from Next-Generation EU (Italian PNRR – M4 C2, Invest 1.3 – D.D. 1551.11-10-2022, PE00000004). CUP MICS D43C22003120001.}
\thanks{The authors are with Politecnico di Milano, Piazza Leonardo da Vinci, 32. Milano (Italy) {\tt\footnotesize marco.faroni@polimi.it}
}
}

\begin{document}

\maketitle

\begin{abstract}
Ensuring human safety in collaborative robotics can compromise efficiency because traditional safety measures increase robot cycle time when human interaction is frequent. 
This paper proposes a safety-aware approach to mitigate efficiency losses without assuming prior knowledge of safety logic. 
Using a deep-learning model, the robot learns the relationship between system state and safety-induced speed reductions based on execution data. 
Our framework does not explicitly predict human motions but directly models the interaction effects on robot speed, simplifying implementation and enhancing generalizability to different safety logics. 
At runtime, the learned model optimizes task selection to minimize cycle time while adhering to safety requirements. 
Experiments on a pick-and-packaging scenario demonstrated significant reductions in cycle times. 
\end{abstract}

\section{Introduction}\label{sec:intro}

\IEEEPARstart{C}{ollaborative} 
robotics is expected to be an enabling technology of the factory of the future, offering benefits such as increased workspace efficiency and the removal of restrictive robot cages. However, robots operating alongside humans in shared workspaces pose significant safety challenges. Ensuring compliance with safety standards (e.g., ISO 10218-2 \cite{ISO10218-2}) is essential to prevent harm, but often results in efficiency trade-offs. 
For instance, commonly used static safety zones cause robots to slow down or stop when a human enters a predefined area, leading to increased cycle times during frequent interactions.

An intuitive way to reduce safety slowdowns is to synchronize the robot and the human’s operations so that the agents avoid entering low-speed regions at the same time. 
Existing methods to address this challenge have focused on optimizing robot scheduling \cite{faccio2023task, SANDRINI2025103006}, human-aware path planning \cite{HAMP,Haddadin:S_star}, or trajectory adjustments based on operator intent \cite{Huang2016, Palleschi2021}. 
While effective, these approaches rely on a priori knowledge of the safety logic, intertwining control and safety design. 
For example, they require that the developer knows how the risk assessment phase will be carried out and, vice versa, that the system integrator can act on the design of the motion planner. 
However, risk assessment and control development typically occur independently and at a different time \cite{VICENTINI2020101921}. 

In practice, obtaining the exact safety configuration is often difficult even after the deployment. 
While in theory one could extract it from risk assessment documents or inspect controller configurations, in reality safety functions are often manufacturer-specific and hidden from the user.
In other cases, the configuration can technically be retrieved, but converting it into a usable set of equations for control purposes requires significant manual effort.
By contrast, learning the relationship between system state and safety-induced slowdowns directly from execution data avoids this burden and provides a more practical and generalizable alternative.

%Moreover, the actual safety configuration is often not readily available even after the cell deployment.
%For example, some safety functionalities are manufacturer-dependent and their actual implementation is not known to the user. 
% when the safety configuration is available from the risk assessment document, it may not be straightforward to convert it into a set of equations.

Motivated by these limitations, we propose a data-driven, safety function-agnostic approach to enhance collaborative robot efficiency by optimizing robot task scheduling without requiring explicit knowledge of safety logic. 
The system uses a deep-learning model to learn the relationship between human-robot interaction modes and robot speed reductions based on process data. 
This data-driven approach has multiple advantages: 
i) It allows integration with existing cells where the safety configuration is inaccessible; 
ii) It adapts naturally to changes in task or layout after retraining, especially considering that training data can be collected during normal operations; 
iii) Unlike model-based methods, which require predicting human motion at future time instants, our approach directly learns the impact of human activity on robot speed.
At runtime, the learned model guides task selection to minimize cycle time. 
To do so, we propose two action selection algorithms that leverage the deep-learning model to minimize expected speed reductions. 
The first algorithm predicts the robot's slowdown for all available actions and chooses the one with the minimum impact on the robot's speed. 
The second algorithm uses a Monte Carlo approach to plan for longer prediction horizons. 

Experiments on a pick\&packaging application showed a significant improvement in the robot cycle time and mitigation of safety speed reductions for both the proposed algorithms.

\section{Related works}\label{subsec: related-works}

Safety standards such as the ISO 10218-2 \cite{ISO10218-2} define frameworks for deploying collaborative robotics systems. 
These standards provide general guidelines for human safety, requiring system integrators to perform case-specific risk analysis and mitigation based on ISO 12100. 
Techniques such as Speed and Separation Monitoring (SSM) and Power and Force Limitation (PFL) ensure safety by applying speed reduction rules. 
However, the speed reduction significantly impacts robot cycle times, leading to inefficiencies in collaboration.

%ISO 10218 \cite{ISO10218-2} and ISO TS 15066 \cite{ISOTS15066} define a framework to deploy safe collaborative applications. 
%Because the variety of collaborative applications is so various, these standards mainly define general requirements that the system integrator has to evaluate case by case according to a risk analysis and mitigation phase (ISO 12100). 
%A common ground defined by ISO 10218 and ISO TS 15066 is that human safety against collisions can be ensured by slowing down the robot when the operator is in a potential collision area. 
%For example, the Speed and Separation Monitoring (SSM) function defines worst-case rules to reduce the robot’s speed always to avoid collisions. Similarly, the Power and Force Limitation (PFL) function limits the robot’s speed to allow safe, harmless contacts. 

%Safety speed reduction causes delays in the robot’s cycle time and may make the collaboration less smooth. 
%Especially when the safety functions are implemented with conservative approaches (e.g., static areas with speed thresholds), the robot often stops as soon as the human enters the shared workspace \cite{SSM:implementation}. 
Previous works addressed this problem by acting on the robot’s control pipeline \cite{Palleschi2021,HAMP,Haddadin:S_star,
faccio2023task,Kanazawa2019,flowers2023spatio,optimal-planning-dynamic-risk-regions}.  %Tonola2021AnytimeIP, PELLEGRINELLI20171

%Among them, some recent works aimed to embed safety specifications into the control problem. 
For example, \cite{HAMP} and \cite{Haddadin:S_star} proposed a safety-aware path planner minimizing the expected robot’s execution time considering also safety slowdowns. 
They define cost functions modeling the SSM and PFL speed reduction rules defined in ISO 10218-2 and search for a path minimizing such cost functions. 
At a lower control level, \cite{pupa:mpc} uses model predictive control to find a feasible trade-off between path following and compliance with safety constraints. 
As for scheduling, \cite{faccio2023task} proposes a scheduler to prevent stops owed to simultaneous access to the same area. 
To do so, they assign penalty terms to each region of the workspace according to ISO 10218-2. 
Other works merge the psycho-physical state of the operator with safety awareness \cite{action-planning-careless,lagomarsino2024pro}. 
These methods translate nominal speed reduction values from ISO 10218-2 into an optimization problem. 
However, ISO 10218-2 defines lower bounds on the human-robot distance, but such values need to be inflated depending on the perception system and the case-specific risk analysis.

Safety-aware methods are tightly coupled with the prediction of the human motion as they shall anticipate the human actions \cite{hierarchical-human-motion-prediction,uncertainty-hamp,scalera2024robust,uncertain-human-predictio-planning}. 
Depending on the length of the prediction horizon, they integrate human prediction in different ways.
Short-term prediction is usually addressed with physics-based models \cite{hermann2015anticipate} or filtering techniques \cite{ferrari2024predicting}. 
Long-term prediction usually leverages additional information on the context and the human goal or use agglomerated estimates of the human occupancy \cite{pellegrinelli2018estimation}. 
Recently, deep-learning approaches have become the most widespread for long-term, context-aware prediction \cite{finean2023motion,long-term-3,toussaint-long-term-4}. 

Our approach differs from prior works in two key aspects. First, it does not assume prior knowledge of safety logic. 
Instead, it learns the correlation between system states and safety-induced speed reductions from process data, enabling generalization across different implementations. 
Second, rather than explicitly modeling the human movements, our approach directly predicts the robot’s slowdown, simplifying the model while providing a meaningful prediction for decision-making. 

\section{Problem Statement}\label{sec:problem}

We consider a robot and an operator working in a shared environment. 
We assume a safety system acting as follows: when the human is not in a cooperative area, the robot moves at a nominal speed; when the human enters a collaborative area, it reduces its speed to a percentage of the nominal speed. 
A speed reduction function computes the speed percentage based on the measured system states. 
For example, common implementations slow the robot down proportionally to the inverse of the human-robot relative distance and velocity or assign fixed values when the human enters pre-defined areas. 

In our setup, the human and the robot take actions asynchronously by choosing online from a pool of available actions. 
Our goal is that the robot selects the action that mitigates the safety speed reduction and, thus, reduces cycle and idle times. 

Let $\stateR \in \StateR$ and $\stateH \in \StateH$ be the robot and human's states, $\actionR \in \ActionR$ the robot's action, where $\StateR$, $\StateH$, $\ActionR$ are the corresponding state and actions spaces.
We denote the safety speed reduction function by $s(\stateR, \stateH) \in [0,1]$. 
This means that the actual robot speed is equal to the nominal speed (i.e., without human presence) multiplied by $s(\stateR,\stateH)$. 
According to the scenario above, we assume (i) $s$ has an unknown shape but its value at time $\bar{t}$ can be observed and (ii) $\stateR$, $\stateH$, $\actionR$ are observable. 
We do not make assumptions about the shape of $s$ (except that it depends on $\stateR$ and $\stateH$).

A robot's action, $\actionR$, consists of moving the robot to a goal state, $\goalR$. 
For this reason, we can define a function that associates $\actionR$ with $\goalR$, such that:
\begin{equation}
\label{eq: get-robot-goal}
    \goalR = \texttt{getRobotGoal}(\actionR) 
\end{equation}
where $\ActionR$ is the set of all robot actions. 
Similarly, we assume the human task is composed of sub-operations, each one associated with a goal position. 
We denote the set of human goals by $\GoalHSet = \{ \GoalHSetI{j} \}$ and the current human goal by $\GoalH$.
To account for human variability, $\GoalHSetI{j}$ is a random variable such that
\begin{equation}
\label{eq: human-distribution}
    \GoalHSetI{j} \sim \mathcal{N}(\mu_{j},\sigma_{j})
\end{equation}
where $\mu_{j}$ and $\sigma_{j}$ are the mean and standard deviation of the Gaussian distribution. 
We assume it is possible to retrieve $\GoalH$ for the current human operation (e.g., through a task segmentation algorithm that identifies the current human operation or explicit feedback on an HMI). 

Our goal is to choose the robot's next action, $\actionR^*$, that minimizes the safety speed reduction intervention. 
This problem can be framed as a Markov Decision Process with an exogenous input (Exo-MDP). 
An Exo-MDP is a particular type of Markov Decision Process (MDP), where the process can be factorized in an endogenous and an exogenous part. 
In our problem, the robot is the endogenous part and the human is the exogenous input. 
In particular, we define the Exo-MDP as a tuple $(\StateR,\ActionR,\StateH,P,R)$ where:
\begin{itemize}
    \item $\StateR$ is the endogenous state space (i.e., the robot position and/or velocity and/or auxiliary variables);
    \item $\ActionR$ is the robot action space;
    \item $\StateH$ is the exogenous input (i.e., the human position and/or velocity)
    \item $P(\stateR'|\stateR,\stateH,\actionR)$ is a probability function that models the transition probability from one endogenous state to another;
    \item $R(\stateR,\stateH)$ is a reward function proportional to the safety speed scaling $s(\stateR,\stateH)$.
\end{itemize}
If the robot selects an action $\actionR(k)$ at a decision step $k$, it will reach the goal $\goalR(k)$ at the next decision step. 
Therefore, we can consider the robot dynamics as deterministic (i.e., $P=1$) and obtain
\begin{equation}
    \stateR(k+1) = \texttt{getRobotGoal}(\actionR(k))
\end{equation}
Note that decision steps are incremental values that correspond to the time instant when the robot ends the ongoing operation and needs to select the next one. 
We can define a function $\tau(k)$ to map decision steps into actual time instants.

The safety scaling factor plays the role of the reward function. 
At decision step $k$, we aim to maximize the safety scaling factor over a time horizon $w$, i.e.,
\begin{equation}
\label{eq: argmax}
    \actionR^*(\tau(k)) =  \argmax_{\actionR \in \ActionR(k)} \mathbb{E}\biggl[\, \int_{\tau(k)}^{\tau(k)+w} s(\stateR(t), \stateH(t)) \,dt \,\biggr]
\end{equation}
where $\ActionR(k)$ is the set of available actions at step $k$. 
Note that $s\in [0,1]$ and the closer is $s$ to 1, the smaller is the robot slowdown. 
However, in our problem, $s$ has an unknown shape. 

Our goal is to learn the correlation between the system's state and the safety effect from data and use it to approximately solve \eqref{eq: argmax} online, every time the robot has to select a new action. 

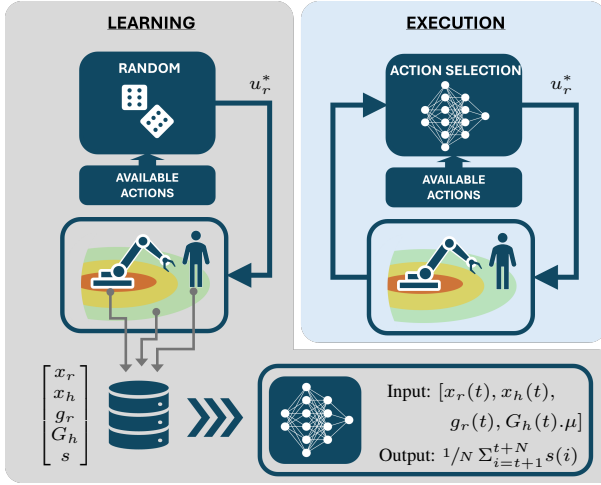
\begin{figure}[tpb]
    \centering
    \begin{tikzpicture}
        \draw (0, 0) node[inner sep=0] {\includegraphics[width=8cm]{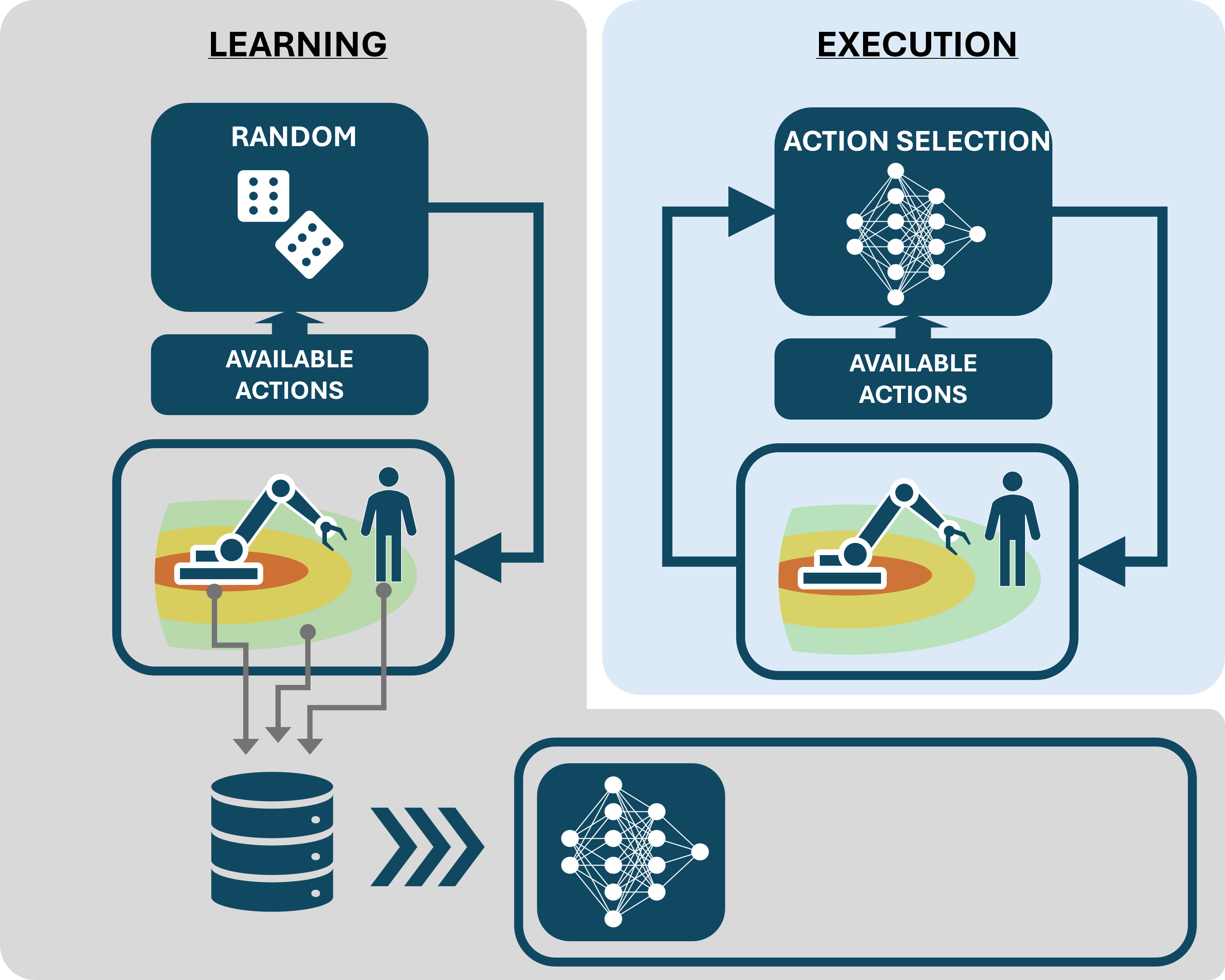}};
        \draw (-0.6, 2.1) node {\scriptsize $\actionR^*$};
        \draw (3.4, 2.1) node {\scriptsize $\actionR^*$};
        \draw (-3.2, -2.3) node {\scriptsize $\begin{bmatrix} \stateR \\ \stateH \\ \goalR \\ \GoalH \\ s \end{bmatrix}$};
        \draw (2.2, -2.0) node {\scriptsize Input: $[\stateR(t), \stateH(t), $};
        \draw (2.8, -2.4) node {\scriptsize $\goalR(t), \GoalH(t).\mu]$};
        \draw (2.3, -2.85) node {\scriptsize Output: $\sfrac{1}{N}\,\Sigma_{i=t+1}^{t+N}s(i)$};
    \end{tikzpicture}
    \caption{Working principle of the proposed approach.}
    \label{fig: working-principle}
\end{figure}

\section{Method}\label{sec:method}

We tackle the problem above in two phases. 
In the learning phase, we collect data from randomized task executions (i.e., let the robot take action randomly) and record $\stateR$, $\stateH$, $\goalR$, $\GoalH$, and $s(\stateR, \stateH)$. 
We use the resulting dataset to train a neural network that predicts the average values of $s$ over a future time window. 

In this work, robot and human's states and goals are Cartesian positions of the robot's end-effector and the human centroid, respectively. 
However, the method can accommodate for different or more complex representations (e.g., position, velocity, and other process variables).

In the execution phase, the robot has to choose its next action among the available ones. 
We propose two approximated methods using the neural network to predict the slowdown associated with each available action and choose the action with the less severe slowdown. 
The working principle of the approach is illustrated in Fig. \ref{fig: working-principle}. 

\subsection{Learning pipeline}\label{sec:learning}

We aim to predict the average value of the scaling function over a given time window. 
We consider the following features: the robot and human's states, $\stateR$, $\stateH$, and the robot and human's goals at the current time, $\goalR$ and $\GoalH$. 
As an example, in our experiments, $\stateR$ and $\stateH$ are the (x,y,z) position of the robot end-effector and the human centroid, $\goalR$ is the final position of the robot trajectory, and $\goalH$ is the mean value of the human task final position.

Our learning architecture uses a feed-forward neural network. 
The loss function is the Mean Squared Error (MSE) of the average scaling factor over $N$ samples, with sampling period $\Delta t$, i.e., $L = || y - \hat{y} ||^2$, where $\hat{s}(t)$ is the observed scaling value at time $t$, $\hat{y}$ is the prediction of the network and
\begin{equation}
\label{eq: loss}
    y = \frac{1}{N+1} \sum_{i=0}^{N} \hat{s}(\bar{t}+i\Delta t).
\end{equation}

While our approach can generalize to any safety function $s$, it is worth analyzing the specific case where $s$ is a staircase function (see Sec. \ref{sec:experiments} for an example). 
As mentioned in Sec. \ref{sec:intro}, this choice is common in industrial applications, where the robot's speed scaling is set to fixed values based on the human position and robot's positions. 
In this case, the safety function can be modeled as
\begin{equation} \label{eq: safety-staircase}
s(x_r,x_h) = \left\{\begin{array}{ll}
        \tilde{s}_1, & \text{if } d(\stateR,\stateH) \in D_1\\
        \tilde{s}_2, & \text{if } d(\stateR,\stateH) \in D_2\\
        \dots  & \\
        \tilde{s}_K, & \text{if } d(\stateR,\stateH) \in D_K
        \end{array}\right.
\end{equation}
where $\tilde{s}_i \in [0,1]$ are constant scaling values, $\tilde{s}_i < \tilde{s}_{i+1}$, $d(\stateR,\stateH)$ is a generic function in the robot and human state, and $D_i$ is a set of states whose membership determines the scaling value. 
The average scaling over $[\bar{t},\bar{t}+N\Delta T]$ is
\begin{equation}
    \bar{s}_{[\bar{t},\bar{t}+N\Delta T]} = \frac{1}{K} \sum_{i=1}^K \alpha_i \tilde{s}_i, \text{ with } \sum_{i=1}^K \alpha_i = 1
\end{equation}
where $\alpha_i$ is the fraction of $[\bar{t},\bar{t}+N\Delta T]$ where $\hat{s}(t)=\tilde{s}_i$. 

Building on this observation, we use a network whose last hidden layer has width equal to $K$ and a Softmax activation function. 
Then, we use a linear output layer and obtain
\begin{equation} \label{eq: safety-function}
    \hat{y} = \beta_0 + \sum_{i=1}^K \beta_i \hat{y}_i'
\end{equation}
where $\hat{y}' \in [0,1]^K$ is the output of the last hidden layer, $\beta \in \mathbb R^{K+1}$ is the weight vector of the output layer, and $\sum_{i=1}^{K} \hat{y}_i'=1$ thanks to the Softmax function. 
The intuition behind this design is that the hidden layers will learn a feature vector $\hat{y}'$ proportional to $\alpha$ while the last hidden layer will learn how to weigh each contribution and yield the average scaling. 

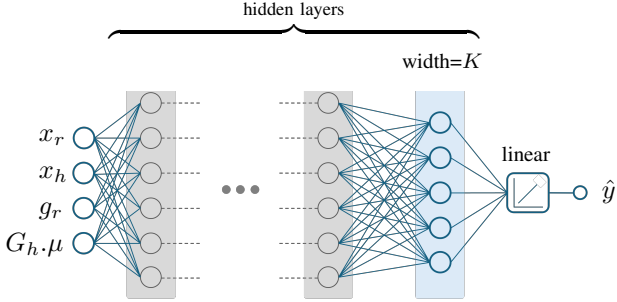
\begin{figure}[tpb]
    \centering
    \setlength{\unitlength}{0.1\columnwidth}
    \begin{picture}(10,5)
        \put(0,0){\includegraphics[trim=3cm 4cm 2cm 5cm, clip, width=\columnwidth]{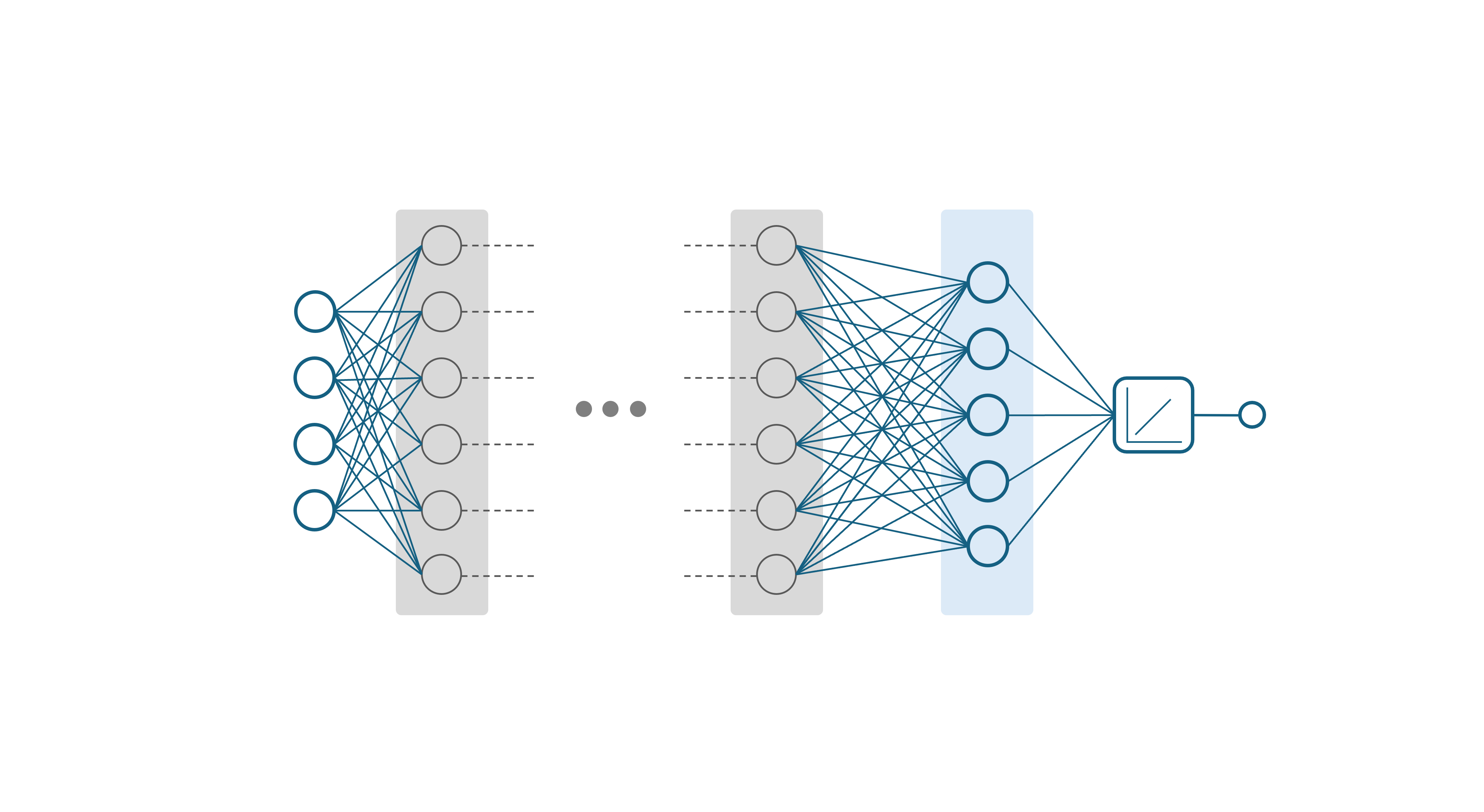}}
        \put(0.75,2.5){$\stateR$}
        \put(0.75,1.95){$\stateH$}
        \put(0.75,1.4){$\goalR$}
        \put(0.25,0.85){$\GoalH.\mu$}
        \put(9.2,1.65){$\hat{y}$}
        \put(1.8,4){$\overbrace{\qquad\qquad\qquad\qquad\qquad\qquad\qquad}^\text{hidden layers}$}
        \put(6.2,3.6){\footnotesize width=$K$}        
        \put(7.7,2.2){\footnotesize linear}        
    \end{picture}
    
    \caption{Neural network architecture.}
    \label{fig: network}
\end{figure}

Because the network does not know $K$ beforehand, we estimate it directly through data clustering as follows:
\begin{enumerate}
    \item We consider the collected values of the scaling factor $\{\hat{s}_i\}$, with $\hat{s}_i\in [0,1]$.
    \item We apply a clustering algorithm on $\{\hat{s}_i\}$ to determine the number of clusters. Most clustering algorithms automatically output the optimal number of clusters. For those that do not (e.g., K-means), it can be computed using the Silhouette or the elbow method \cite{Silhouettes}. 
    \item We approximate the number of steps $K$ as equal to the number of clusters.
\end{enumerate}
Finally, we set the width of the last hidden layer equal to $K$. 
The resulting network is exemplified in Fig. \ref{fig: network} for $K$=5.

This approach works only if the safety function is a staircase function. 
Otherwise, the learning pipeline may benefit from adjustments to the last hidden layers. 

\subsection{Execution pipeline}\label{sec:execution}

\begin{algorithm}[tpb]
\caption{Greedy Action Selection Algorithm}
\label{alg: greedy}
\small

\SetKwFunction{FGetAvailableActions}{getAvailableActions}
\SetKwFunction{FGetRobotGoal}{getRobotGoal}

\KwIn{Observed states $\stateR$ and $\stateH$, Current human goal $\GoalH$, Neural Network \textit{NN}}
\KwOut{Selected action \textit{best\_action}}

\SetKwFunction{FMain}{SelectAction}
\SetKwFunction{FPredict}{predict}

best\_action $\gets$ \texttt{None}\;
highest\_reward $\gets$ $-\infty$\;

actions $\gets$ \FGetAvailableActions($\stateR$)\;
\For{each action in actions}{ \label{alg: alg1-loop}
    $\goalR\gets$ \FGetRobotGoal(action)\;
    reward $\gets$ NN.\FPredict{$\stateR$,$\stateH$,$\goalR$,$\GoalH.\mu$}\; \label{alg: alg1-predict}
    \If{reward > highest\_reward}{
        best\_action $\gets$ action\;
        highest\_reward $\gets$ reward\;
    }
}

\textbf{Return} best\_action\;
\end{algorithm}

We propose two selection algorithms to choose the most convenient action based on the neural network's predicted scaling values. 
The first algorithm is a greedy selection strategy that works well for short time horizons, while the second one leverages the Exo-MDP formulation in a Monte Carlo Planning framework.

\subsubsection{Greedy Action Selection}
Given a desired time window, $w$, we set $N=w/\Delta t$ in \eqref{eq: loss}. 
Because the network predicts the average value over the time interval $(\bar{t},\bar{t} + N \Delta t]$, we can solve \eqref{eq: argmax} by simply inferring the predicted scaling factor, $\hat{y}$, for all available actions and selecting the one with the highest value (i.e., the one that introduces the smallest average speed reduction):
\begin{equation}
    \actionR^* = \argmax_{\actionR \in \ActionR(\bar{t})}\, \hat{y}\big(\stateR(\bar{t}), \stateH(\bar{t}), \goalR(\bar{t}), \GoalH(\bar{t}).\mu\big)
\end{equation}
where $\goalR(\bar{t}) = \texttt{getRobotGoal}(\actionR(\bar{t}))$ and $\GoalH(\bar{t}).\mu$ is the mean value of the current human goal distribution.

A pseudo-code of the procedure is in Alg. \ref{alg: greedy}. 
At each decision time, we predict the expected safety scaling (line \ref{alg: alg1-predict}) and return the action associated with the highest value. 
Note that the loop at line \ref{alg: alg1-loop} can be easily parallelized for computational efficiency.

We refer to this strategy as greedy because it only considers the best action over a short horizon, $w$. 
In our experiments, $w$ is approximately equal to the average duration of one action.

\subsubsection{Monte Carlo Action Selection}
The Greedy Action Selection does not work well for long horizons because the scaling prediction worsens when the time horizon encompasses several robot actions and human goals. 
In general, Exo-MDPs can be solved with Monte Carlo search methods like Monte Carlo Tree Search or Monte Carlo Planning. 

To suit computational time requirements, we propose an anytime search algorithm inspired by Monte Carlo Planning. 
The algorithm works as follows (see Alg. \ref{alg: monte-carlo}). 
Starting from the current state, it expands all children by applying each available robot's action. 
Then, it starts a thread for each child and applies random rollouts (see Alg. \ref{alg: performRollout}) until the sequence reaches a terminal state. 
In our case, a terminal state is either a state where no available actions exist or a maximum length of the sequence is reached. 
Each rollout returns a reward associated with one of the children. 
After the timeout, the algorithm computes the average reward for each child and selects the action that generated the highest reward child. 

The difficulty of applying Monte Carlo methods to Exo-MDPs is that the exogenous input's dynamics is unknown. 
For this reason, it is not possible to propagate the exogenous state at a generic iteration $i$. 
A solution is to sample it from a probability distribution. 
In our problem, this applies both to the human state and goal.
To sample meaningful values of $\stateH(i)$ and $\goalH(i)$, we select a goal $\GoalH(i) = \GoalHSetI{j}$ where
\begin{equation}
    \GoalHSetI{j} \sim \mathcal{U}(\GoalHSet),
\end{equation}
and $\mathcal{U}$ is the uniform distribution over all possible goals.
Considering that, for the $j$th goal, $\GoalHSetI{j} \sim \mathcal{N}(\mu_{j},\sigma_{j})$, we choose
\begin{equation}
    \stateH(i) \sim \GoalHSetI{j} \text{ and } \goalH(i) = \mu_j.
\end{equation}
This allows us to obtain concrete values of $\stateH$ and $\goalH$ to propagate the state over the tree and estimate the reward at future steps\footnote{This approximation relies on the assumption that the operator will spend most of the time near their goal position. 
In general, $\stateH$ can be sampled from an occupancy map conditioned on the chosen goal, $\GoalH(i)$.}.
The resulting propagation function is shown in Alg. \ref{alg: propagate} and forms the basis of the random rollouts described in Alg. \ref{alg: performRollout}). 

Alg. \ref{alg: monte-carlo} is a simplified version of Monte Carlo Planning that allows for straightforward parallelization and a faster computation compatible with the intended online implementation. 
On the other hand, the search is shallower than standard MCP because it uses random rollouts instead of the iterative exploration-v.-exploitation trade-off typical of MCP.

\begin{algorithm}[t]
\caption{Monte Carlo Selection Algorithm}
\label{alg: monte-carlo}
\small
\KwIn{Observed states $\stateR$ and $\stateH$, Observed human goal $\GoalH$, Neural Network \textit{NN}, Discount factor $\gamma$}
\KwOut{Selected action \textit{best\_action}}

\SetKwFunction{FRandom}{random}
\SetKwFunction{FPerformRollout}{performRollout}
\SetKwFunction{FGetAvailableActions}{getAvailableActions}
\SetKwFunction{FGetResult}{getThreadResult}
\SetKwFunction{FPropagate}{propagate}
\SetKwFunction{FIsTerminal}{isTerminalState}

\SetKw{KwTo}{to}

actions $\gets$ \FGetAvailableActions{$\stateR$}\;
best\_action $\gets$ \texttt{None}; \,
best\_average\_reward $\gets$ $-\infty$\;

\For{each action in actions}{
    \textbf{Start a new thread}\;
    \texttt{Threaded function} evaluateAction(action, max\_time)\;
    \Begin{
        $\stateR$, total\_reward $\gets$ \FPropagate($\stateR$, $\stateH$, $\GoalH$, action)\;
        i $\gets$ 0\;
        \While{elapsed\_time < max\_time}{
            reward $\gets$ \FPerformRollout{$\stateR$}\;
            total\_reward $\gets$ total\_reward + $\gamma^i$ reward\;
            i $\gets$ $i$ + 1\;
        }
        \textbf{Return} total\_reward / iterations \;
    }
}

\For{each action thread in active threads}{
    \textbf{Wait for all threads to complete}\;
    average\_reward $\gets$ \FGetResult{action}\;
    \If{average\_reward > best\_average\_reward}{
        best\_average\_reward $\gets$ average\_reward\;
        best\_action $\gets$ action\;
    }
}

\textbf{Return} best\_action\;

\end{algorithm}

\begin{algorithm}[t]
\caption{\texttt{performRollout}}
\label{alg: performRollout}

\small

\KwIn{$\stateR$}
\KwOut{\textit{total\_reward}}

\SetKwFunction{FMain}{performRollout}

total\_reward $\gets$ 0\;
\While{\FIsTerminal{$\stateR$} is false}{
    actions $\gets$ \FGetAvailableActions{$\stateR$}\;
    action $\gets$ \FRandom{actions}\;
    $\GoalH$ $\gets$ \FRandom{$\GoalHSet$}\; \label{alg: alg3-rnd-goal}
    $\stateR$, reward $\gets$ \FPropagate($\stateR$, $\GoalH$, action)\; \label{alg: alg3-propagate}
    total\_reward $\gets$ total\_reward + reward\;
}
\textbf{Return} total\_reward\;
\end{algorithm}

\begin{algorithm}[b]
\caption{\texttt{propagate}}
\label{alg: propagate}
\small

\KwIn{$\stateR$, $\GoalH$, \textit{action}}
\KwOut{$\stateR$, \textit{reward}}

\SetKwFunction{FMain}{propagate}
    $\goalR$ $\gets$ \FGetRobotGoal{action} \; \label{alg: alg4-propagated-state}
    $\stateH \sim \GoalH$\;
    reward $\gets$ NN.\FPredict{$\stateR$, $\stateH$, $\goalR$, $\GoalH.\mu$}\; \label{alg: alg4-nn}
    $\stateR$ $\gets$ $\goalR$\;
\textbf{Return} $\stateR$, reward\;
\end{algorithm}

\section{Experiments}\label{sec:experiments}

We demonstrate our approach in a simulated pick\&packaging example and a real-world scenario. 
Simulations evaluate the learning pipeline and provide an ablation study of the scheduling approach.
The real-world scenario proves the effectiveness of the approach in execution against several baselines.

\subsection{Simulated pick\&packaging scenario}
\label{sec: simulations}

\begin{figure}[tpb]
    \centering
    \includegraphics[trim={4cm 0cm 2cm 1.5cm}, clip, width=0.82\columnwidth]{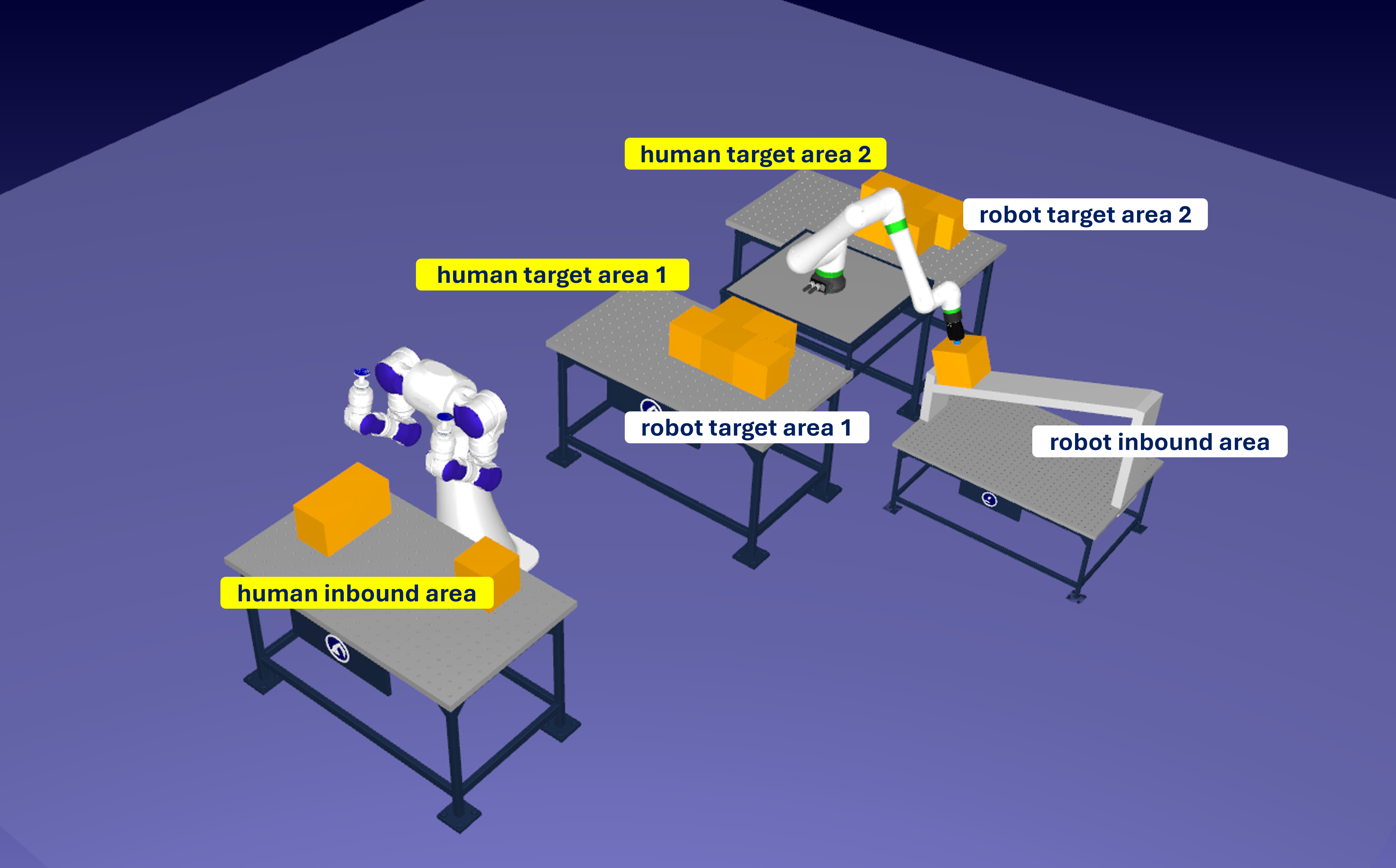} \,
    \caption{Simulation scenario.}
    \label{fig: simulation-scenario}
\end{figure}

We consider a box-picking scenario with a robot and an operator in the RoboDK simulator. 
As RoboDK does not have a human simulator, we mimic the operator by using a humanoid robot, Motoman SDA10F, equipped with an omnidirectional mobile base. 
The manipulator is a 6-dof Fanuc CRX10iA with a vacuum gripper. 
The scenario is in Fig. \ref{fig: simulation-scenario}.

\begin{figure}[tpb]
    \centering
    \begin{tikzpicture}
        \begin{axis}[
            width=0.8\columnwidth,
            height=0.43\columnwidth,
            axis lines=middle,
            xlabel style={at={(axis description cs:0.5,-0.25)}, anchor=north},
            xlabel={human-robot distance},
            ylabel={$s$},
            xtick={1,2,3,4},
            xticklabels={$d_1$, $d_2$, $d_3$, $d_4$},
            ytick={0.25, 0.5, 0.75, 1},
            ymin=0, ymax=1.2,
            xmin=0, xmax=5,
            domain=0:4,
            samples=200,
            grid=both,
            grid style={dashed, gray!30}
        ]
        % Draw the staircase function
        \addplot[
            blue,
            thick,
            domain=0:1
        ] {0};
        \addplot[
            blue,
            thick,
            domain=1:2
        ] {0.25};
        \addplot[
            blue,
            thick,
            domain=2:3
        ] {0.5};
        \addplot[
            blue,
            thick,
            domain=3:4
        ] {0.75};
        \addplot[
            blue,
            thick,
            domain=4:5
        ] {1};
        \end{axis}
    \end{tikzpicture}
    \caption{Example of safety speed scaling function for $K=5$.}
    \label{fig: safety-function}
    \vspace{-0.4cm}
\end{figure}
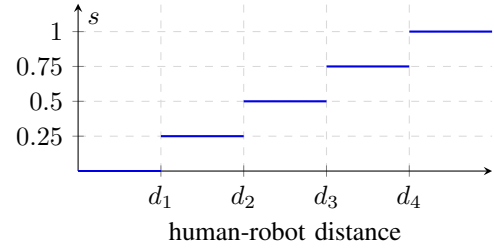

The robot and the operator pick and place boxes from inbound to outbound areas. 
The robot's inbound area consists of a conveyor, while the outbound areas are on the two tables on the robot's sides and accommodate up to 5 boxes each. 
The human inbound area contains 6 boxes, while the outbound areas are near the robot and contain 3 boxes each. 
The robot slows down according to a stepwise safety function, which can be modeled according to \eqref{eq: safety-staircase} by choosing $d=||x_r-x_h||$,
$D_i = [d_{i-1}, d_i)$, $d_0 = 0$, $d_K = +\infty$, and $\{d_i\}$ evenly spaced for $i=1,...,K\!-\!1$.
An example for $K=5$ is in Fig. \ref{fig: safety-function}. 
In the experiments, we used $d_4 = 2$ m.

Every time the robot picks a new box, we aim to estimate the expected future slowdown for all available placing points. 
A video of the process is available in the multimedia attachments. 

\subsubsection{Data Collection and Training}
\label{sec: sim-data-collection}

We run the process 1000 times and, with a frequency of 10 Hz, record the following data for each time $\bar{t}$:
\begin{itemize}
    \item the robot's end-effector position, $\stateR(\bar{t}) \in \mathbb R^3$;
    \item the human's centroid position, $\stateH(\bar{t}) \in \mathbb R^3$;
    \item the robot's goal position, $\goalR(\bar{t}) \in \mathbb R^3$;
    \item the human's goal position, $\GoalH(\bar{t}) \in \mathbb R^3$;
    \item the robot's speed scaling, $s(\bar{t}) \in [0,1]$.
\end{itemize}
To account for the human movements' stochasticity, we randomize the path of the operator: the human's goal is sampled from a Gaussian distribution centered in the nominal goal position with a standard deviation of 0.05 m in the horizontal plane; the midpoint of the path is sampled from a Gaussian distribution centered in a nominal midpoint and with standard deviation equal to 0.25 m in the horizontal plane.

We use an 80\%-20\% training/test split of the collected dataset and we train a feed-forward neural network. 
The design of the network follows the method described in Sec. \ref{sec:learning} for staircase safety functions.
The network has 12 input features, the hidden layers' width is equal to 64 (using batch normalization and ReLU activation), and a last hidden layer with a width equal to $K$ and a Softmax activation function. 
The output layer is linear and yields a scalar. 

\subsection{Learning Results}
\begin{figure}[tpb]
    \centering
    \subfloat[][Simulations]
	{\includegraphics[trim={11.5cm 0cm 0cm 1.0cm}, clip, width=0.495\columnwidth]{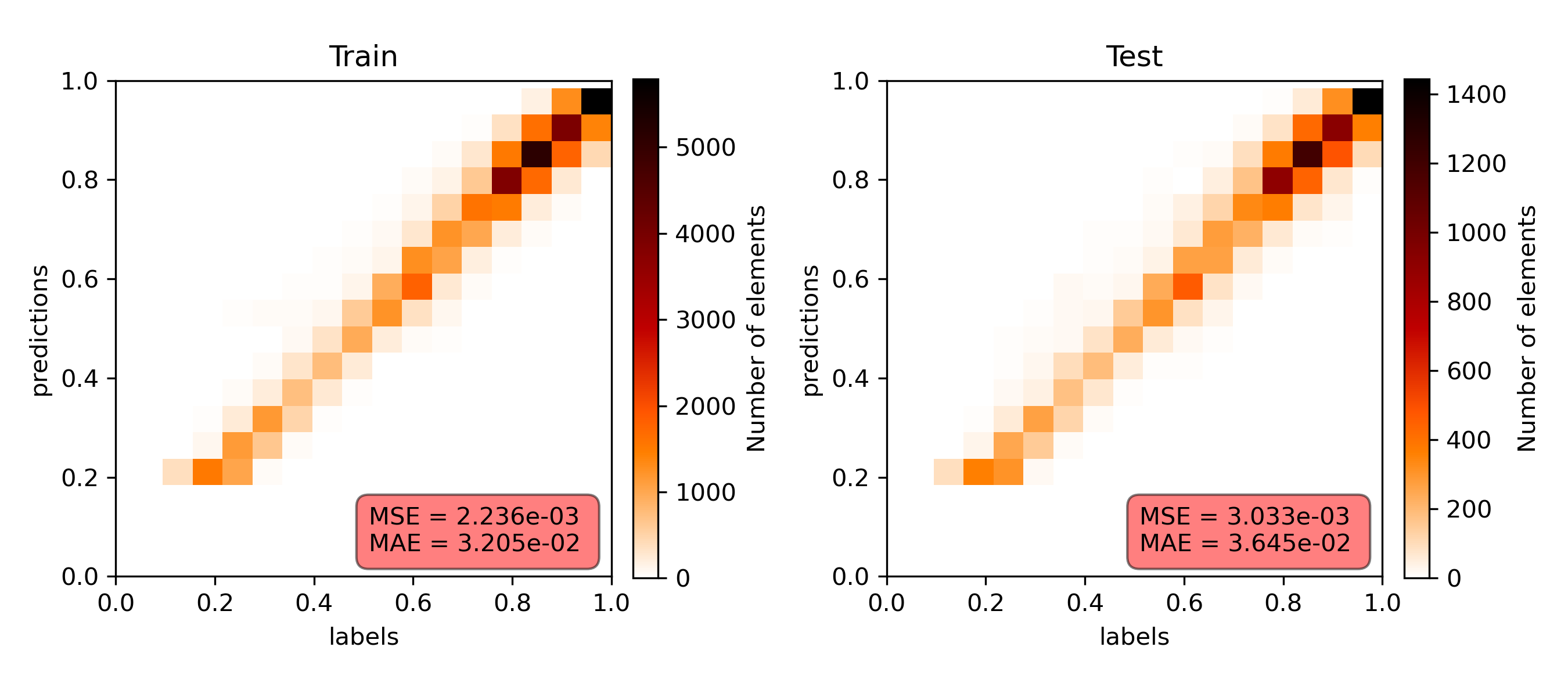}     \label{fig: mse-simulation}}
	\subfloat[][Real-world]
	{\includegraphics[trim={11.5cm 0cm 0cm 0.9cm}, clip, width=0.49\columnwidth]{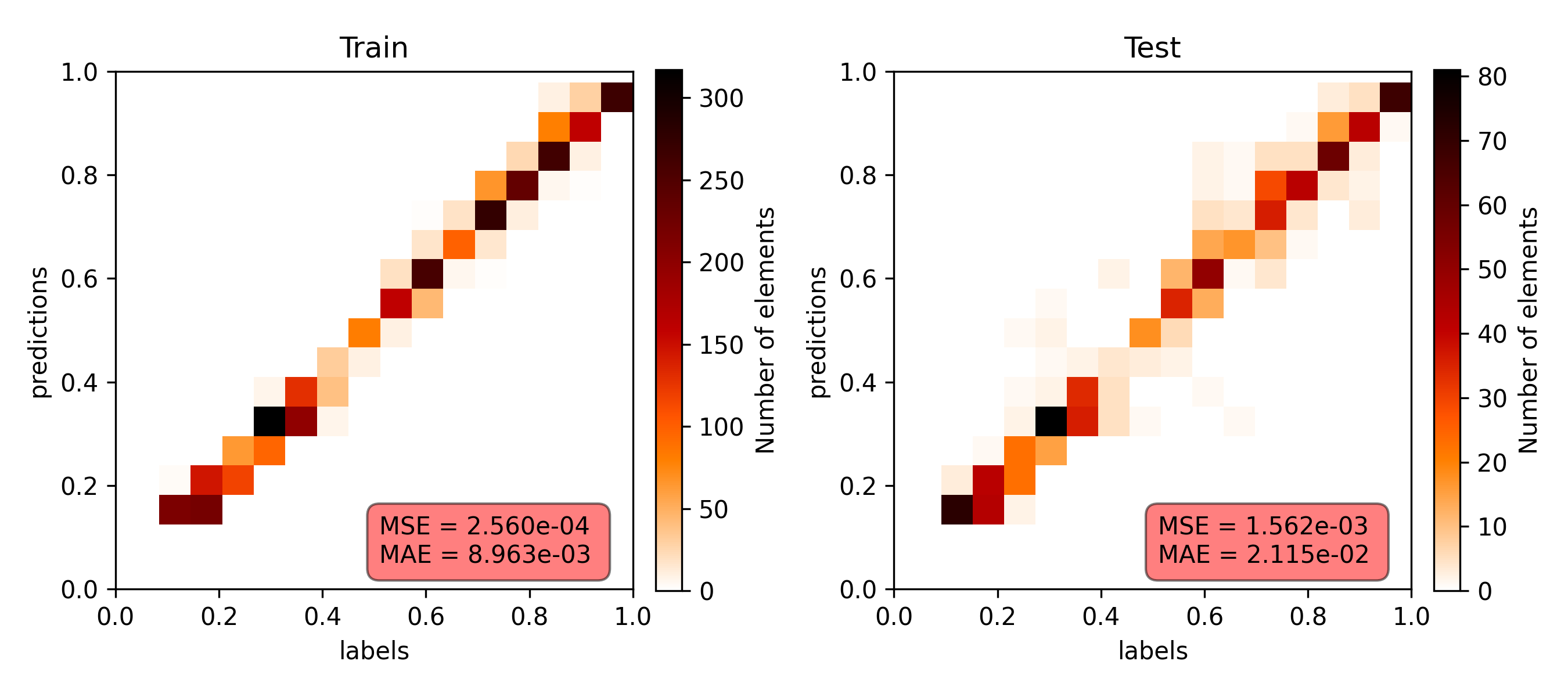}     \label{fig: mse-experiments}}
    \caption{Actual v. predicted average scaling values on test datasets. The heatmap represents the density of samples. A high density along the diagonal denotes a high regression accuracy.}
    \label{fig: mse-heatmap}
\end{figure}

Fig. \ref{fig: mse-simulation} shows the accuracy of the network on the test sets with $K=5 $ and a predictive horizon of 14 seconds (equal to the average of robot tasks' durations estimated during the data collection). 
Similar accuracy values were obtained for predictive horizons in the range of 8--20 seconds. 
The diagonal pattern and low MSE values clearly show the good predictive capabilities of the regressor.
To demonstrate that generality of the approach across safety configurations, we evaluate the prediction accuracy for different values of $K$. 
Note that the design principles of the network remain the same for different values of $K$, as described in Section \ref{sec:learning}. 
The only hyper-parameter is the number of hidden layers, which may increase for complicated functions.
Table \ref{tab: varying-K} shows the accuracy and the number of hidden layers of the network for $K\in \{3,5,10,20\}$.
The results show that the accuracy is consistent for different values of $K$.
As expected, large values of $K$ may benefit from additional hidden layers to achieve a satisfactory accuracy.
It is worth mentioning that the choice of the number of hidden layers can be easily automatized through grid search.

\begin{table}[tpb]
    \caption{Learning results for different values of $K$.}
    \centering
    \begin{tabular}{ccccc}
       \toprule
       $K$  &  3 & 5 & 10 & 20 \\
       \midrule
       MSE $\cdot 10^3$  & 3.22 & 3.03 & 4.07 & 5.45\\
       \# hidden layers & 5 & 5 & 6 & 6 \\
       \bottomrule
    \end{tabular}
    \vspace{-0.3cm}
    \label{tab: varying-K}
\end{table}

\begin{table}[t]
    \caption{Simulation results. Average values of the execution time (per task) and speed scaling for all methods. Values in parenthesis refer to std. deviation.}
    \centering
    \begin{tabular}{lcc}
        \toprule
                        & exec. time & scaling \\
        \midrule
        \footnotesize\texttt{greedy} & \textbf{17.7}(2.1) & \textbf{0.82} \\
        \footnotesize\texttt{greedy-inacc-1} & 18.7(3.2) & 0.78\\
        \footnotesize\texttt{greedy-inacc-2} & 19.3(2.7) & 0.75\\
        \footnotesize\texttt{random} & 20.1(3.1) & 0.72\\
        \bottomrule
    \end{tabular}
    \label{tab: sim-results}
\end{table}

\subsection{Execution Results}
The main motivation behind this work is that the knowledge of the safety function is often uncertain or hard to model explicitly.
For this reason, we evaluate the performance of a selection strategy when the safety function is wrongly modeled. 
To this end, we consider a test scenario with $K=5$ and compare the following cases:
\begin{itemize}
    \item[]\!\!\!\!\texttt{greedy}: our greedy action selection method (Sec. \ref{sec:execution}).
    \item[]\!\!\!\!\texttt{greedy-inacc-1}: same greedy strategy, but the scaling predictor is trained on a scenario with $K=3$.
    \item[]\!\!\!\!\texttt{greedy-inacc-2}: same greedy strategy, but the scaling predictor is trained on a scenario with inflated distance thresholds $d_i^*$, such that $d_i^*=1.2 \, d_i$ for $i=1,...4$. 
\end{itemize}
We also consider a \texttt{random} selection strategy, where the next robot's action is randomly chosen among the available ones.

Table \ref{tab: sim-results} shows the average execution time and safety speed slowdown for all the methods.
\texttt{greedy} reduced the execution time by around 12\% with respect to \texttt{random}.
By contrast, \texttt{greedy-inacc-1} and \texttt{greedy-inacc-2} respectively increased execution time by 6\% and 9\% compared to \texttt{greedy}, showing that inaccurate knowledge of the safety function significantly degrades task selection performance.

\subsection{Real scenario}
\label{sec: exp-scenario}

We consider a pick\&packaging scenario with a human operator and a collaborative robot, Universal Robots UR5e. 
The robot picks small boxes from a slider and places them into one of the four boxes on the table. 
In the meantime, the operator performs operations such as packing items, inspecting the packages, and refilling items. 
Thus, the robot has one picking point and four placing areas, while the human moves among three areas spanning both the shared and the non-shared workspaces. 
The robot slows down according to the stepwise function shown in Fig. \ref{fig: safety-function} with $d_1$=0.6 m, $d_2$=0.8 m, $d_3$=0.1.2 m, and $d_4$=1.6 m. 
The scenario is shown in Fig. \ref{fig: exp-video}.

We tested two variants of the scenario. 
\begin{itemize}
    \item In the \textbf{first variant}, each box is replaced as soon as it is full to emulate a continuous flow use case. 
    From the action selection point of view, this means that the robot can always choose any box.
    \item In the \textbf{second variant}, the boxes are replaced only when all of them are full. 
    This means that the robot's available actions decrease according to the previous choices.
\end{itemize}

\subsubsection{Data Collection and Training}
\label{sec: exp-data-collection}

Similarly to the simulations, we collect the human's centroid and goal, the robot's end-effector and goal, and the speed scaling value. 
The robot position is computed by applying forward kinematics to the robot's measured joint state; the robot's goal is known as it is decided by the action selection algorithm; the robot's speed scaling is available from the robot's APIs.
The operator's position is estimated by a skeleton tracking algorithm using RGB-D images from a fixed Intel Realsense d435i camera. 
As for the operator's goal, we approximate it with the average human's position during an operation and a standard deviation equal to 0.1 m.

Note that the order of the operations is not fixed as the operator selects its task on the fly. 
To identify the ongoing human operation (and the associated goal, $\GoalH$), the operator presses a button when it starts a new operation. 

The neural network structure is the same as in Sec. \ref{sec: sim-data-collection}. 
To train it, we collect data for around 90 minutes with a sampling rate of 10 Hz. 
Fig. \ref{fig: mse-experiments} shows the regression accuracy, which is consistent with the simulation results. 

\subsubsection{Metrics and Baselines}
\label{sec: exp-matrics}

We aim to demonstrate that our approach reduces the cycle time and safety speed slowdowns. 
Thus, we measure the robot's execution time for each time and the average speed scaling during execution. 

We compare the following action selection algorithms:
\begin{itemize}
    \item[]\!\!\!\!\texttt{random}: the next robot's action is randomly chosen among the available ones.
    \item[]\!\!\!\!\texttt{round-robin}: the robot selects the next placing box in sequential order from the first to the fourth box. Full boxes are skipped.
    \item[]\!\!\!\!\texttt{reactive}: the robot selects the box furthest from the human position at the current time.
    \item[]\!\!\!\!\texttt{greedy}: our greedy action selection method (Sec. \ref{sec:execution}).
    \item[]\!\!\!\!\texttt{monte carlo}: our MCP variant (Sec. \ref{sec:execution}).
\end{itemize}

\subsubsection{Implementation notes}\label{sec: exp-implementation}
The main tunable parameter in our method is the predictive window, $w$. 
\texttt{greedy} and \texttt{monte carlo} use the neural network to predict the average scaling value over the next robot's action. 
For this reason, we set $w$ equal to the average duration of the robot's actions ($\sim$14 s) in the \texttt{greedy} method. 
While \texttt{greedy} has a negligible computational delay, \texttt{monte carlo} requires a large number of rollouts to find a good solution. 
As the robot picking task starts and ends at the same point, we run \texttt{monte carlo} in masked time during the picking operation. 
This strategy grants $\sim$4 seconds to find a solution but introduces a mismatch in $\stateH$.
As a matter of fact, the value of $\stateH$ when the robot applies the action is different from that used by the algorithm. 
To account for this mismatch, we use $w$=18 s in \texttt{monte carlo} to account for the picking phase duration. 

\begin{table}[tpb]
    \caption{Experimental results. Average values of the execution time (per task) and speed scaling for all methods. Values in parenthesis refer to std. deviation.}
    \centering
    \begin{tabular}{lcccc}
        \toprule
                        & \multicolumn{2}{c}{variant 1} & \multicolumn{2}{c}{variant 2} \\
        \midrule
                        & exec. time & scaling & exec. time &  scaling \\
        \midrule
        \footnotesize\texttt{random} & 24.7(6.6) & 0.60 & 24.9(6.6) & 0.59 \\
        \footnotesize\texttt{round-robin} & 26.8(5.9) & 0.55 & 27.1(6.6) & 0.54 \\
        \footnotesize\texttt{reactive} & 24.5(4.7) & 0.60 & 25.1(5.5) & 0.59 \\
        \footnotesize\texttt{greedy} & \textbf{18.6}(1.7) & \textbf{0.74} & \textbf{20.2}(3.9) & \textbf{0.68} \\
        \footnotesize\texttt{monte carlo} & 20.3(3.6) & 0.72 & 20.7(3.9) & 0.66 \\
        \bottomrule
    \end{tabular}
    \label{tab: exp-results}
    \vspace{-0.3cm}
\end{table}

\begin{figure*}[tpb]
    \centering
    \includegraphics[width=0.95\textwidth]{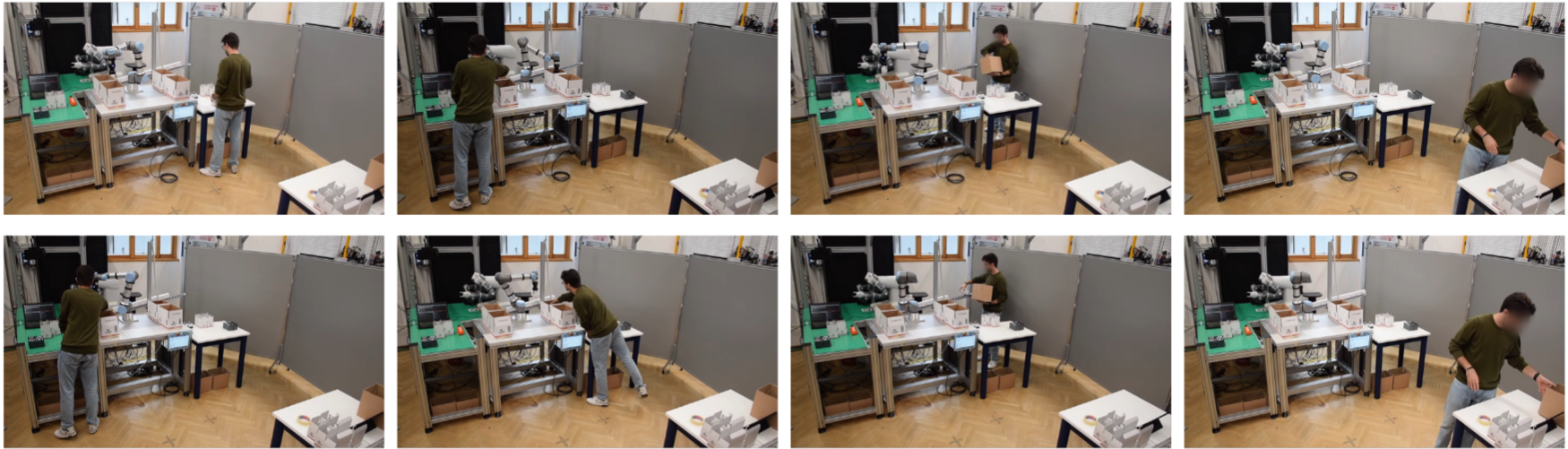}
    \caption{Screenshots from videos of the experiments. (top) \texttt{greedy}. (bottom) \texttt{random}.}
    \label{fig: exp-video}
    \vspace{-0.4cm}
\end{figure*}

\begin{figure}[tpb]
    \centering
    \includegraphics[width=0.99\columnwidth]{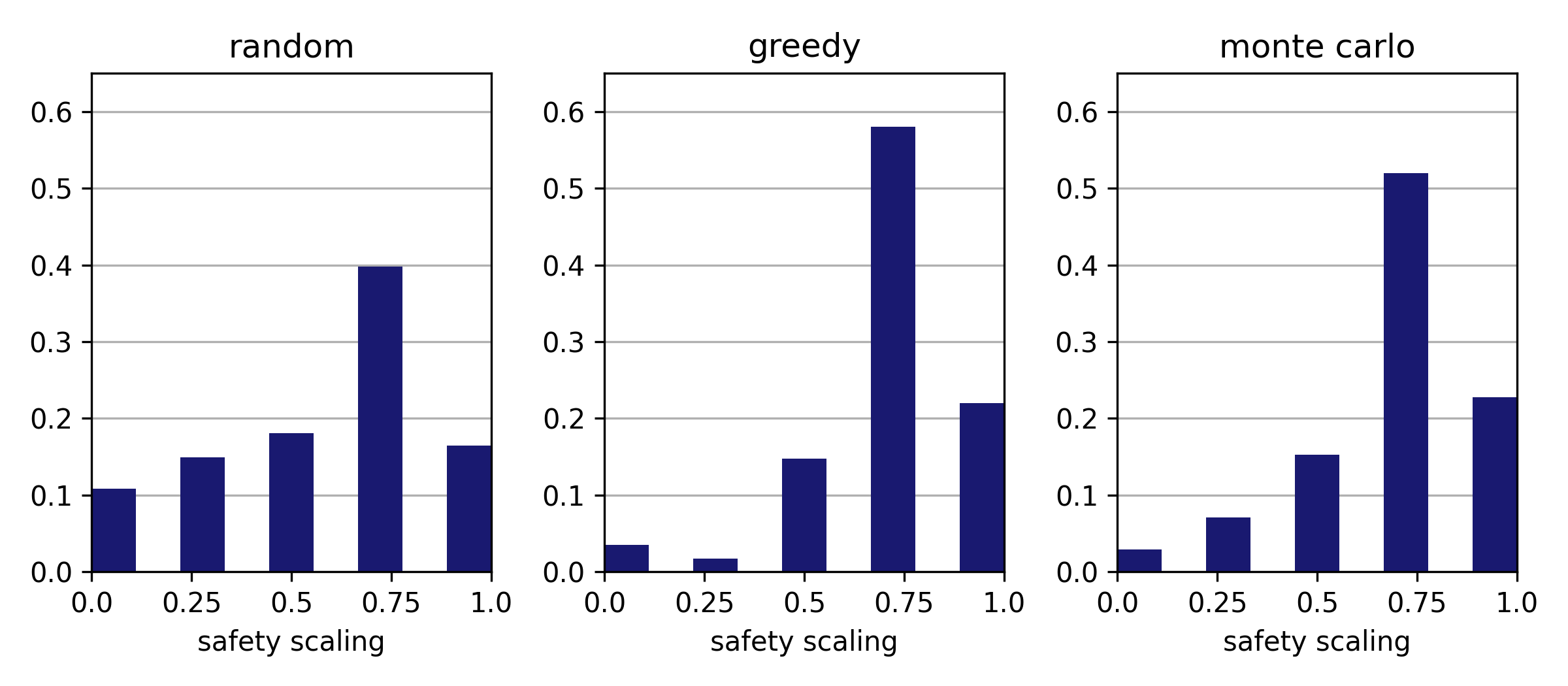}
    \caption{Experimental results. Statistic distribution of the speed scaling factor in the second scenario variant.}
    \label{fig: exp-scaling-results}
    \vspace{-0.4cm}
\end{figure}

\subsubsection{Results}\label{sec: exp-results}

For the first variant, we run one test of around 25 minutes for each method. 
This corresponds to a number of action selections between 60 and 80 for each method. 
As for the second variant, we run the whole process 15 times for each method, for a total of 180 action selections for each method. 

Table \ref{tab: exp-results} summarizes the results. 
In the first variant, \texttt{greedy} gives the best results, followed by \texttt{monte carlo}. 
Compared to \texttt{random}, they improve the robot execution time of about -25\% and -18\% and the average speed scaling of about +23\% and +20\%, respectively. 
The second variant yields similar results: \texttt{greedy} and \texttt{monte carlo} outperform the baseline of about -19\% and -17\% in terms of execution time and of about +15\% and +11\% in terms of average speed scaling. 
The proposed methods also show a smaller variance.

As an example, Fig. \ref{fig: exp-scaling-results} compares the distribution of the speed scaling during one execution of the second variant. 
For the sake of brevity, we only show \texttt{random}, \texttt{greedy}, and \texttt{monte carlo}. 
The histograms show that \texttt{greedy} and \texttt{monte carlo} lead to a higher frequency of high scaling values (i.e., $s$=0.75 and $s$=1.0), which means these methods correctly choose the next action in such a way as to reduce the impact of the safety speed reduction.

We did not observe a significant difference between the \texttt{random} selection and the other baselines, except that \texttt{reactive} shows a smaller variance. 
Although \texttt{reactive} is human-aware, this strategy is unable to choose convenient actions because it is based only on the current operator's position. 
This strategy is ineffective because the scenario requires the operator to move frequently across the workspace. 

In our experiments, \texttt{greedy} and \texttt{monte carlo} gave comparable results. 
One reason for this is that the scenario did not involve a strong coupling between current and future actions' rewards, which leads to good results for the greedy algorithm. 
Moreover, \texttt{monte carlo} performance is affected by implementation issues. 
First, the algorithm is triggered in advance as described in Sec. \ref{sec: exp-implementation}, meaning that the operator may be in a different position than expected at planning time; ii) 
Second, \texttt{monte carlo} is often unable to find optimal solutions in the allotted time. 
Improvements on the \texttt{monte carlo} implementation will be object of future works.

Finally, Fig. \ref{fig: exp-video} shows snapshots taken from an execution using the \texttt{greedy} and the \texttt{random}  methods. 
The pictures show how the proposed method (top pictures in Fig. \ref{fig: exp-video}) picks actions that do not interfere with the operator by anticipating human movements. 
On the contrary, the \texttt{random} method (bottom pictures in Fig. \ref{fig: exp-video}) often drives the robot close to the region occupied by the human (see the attached video).

\section{Conclusions}\label{sec:conclusions}

This paper presented a safety function-agnostic method for optimizing collaborative robot task selection by learning the correlation between system state and safety-induced speed reductions. 
One limitation of the learning-based approach is that it assumes a level of repeatability in human executions, making it best suited for structured tasks. 
Future work will focus on extending the method’s capabilities to partially unstructured tasks by incorporating context-aware models and using automatic task segmentation to infer the human current operation without explicit feedback.

\bibliographystyle{IEEEtran}
\bibliography{bib,bib_new}

\end{document}